\documentclass[runningheads]{llncs}
\usepackage[T1]{fontenc}
\usepackage{graphicx}
\usepackage{amsmath}
\usepackage{amssymb}
\usepackage{amsfonts}
\usepackage{subcaption}
\usepackage{booktabs}
\usepackage{multirow}
\usepackage{placeins}
\usepackage{float}      
\usepackage{placeins} 
\usepackage{hyperref}

\begin{document}
\title{Usefulness of Quantile-Aware Diffusion Modeling for Highly Imbalanced Tabular Data%
\thanks{Accepted for publication in Springer Lecture Notes in Computer Science (LNCS), CSoNet 2026.}}
%
%
\author{Abu Talha\inst{1}\orcidID{0009-0003-9667-667X} \and
Peng Liu\inst{2}\orcidID{0000-0002-0492-0029} \and
Souradyuti Paul\inst{1}\orcidID{0009-0008-1361-3072}}
%
\authorrunning{A. Talha et al.}
%
\institute{Indian Institute of Technology Bhilai, Durg, C.G. 491002, India \and
Loughborough University, Loughborough, LE11 3TU, United Kingdom
\email{P.Liu4@lboro.ac.uk}\\
\email{\{abut,souradyuti\}@iitbhilai.ac.in}}
\maketitle              
\begin{abstract}

Classification problem in the context of highly imbalanced data is a major challenge in many real-world applications (e.g., FinTech, healthcare, etc.). In these cases, the vast majority of instances belong to a single class and a small fraction represent the minority class (often the most critical class). Recently, diffusion models have emerged as powerful approaches to reduce the degree of ``imbalanced-ness'' in the dataset; they work by generating synthetic data by capturing complex data distributions using iterative transformations. However, standard diffusion models are not inherently suited to highly skewed or heavy-tailed data, due to inbuilt quadratic error loss, which lacks the structural sensitivity to capture rare, extreme values, and minority-class nuances. We propose a \textit{novel} approach, namely, Quantile-TabDDPM, based on a quantile-regularized denoising objective that combines the standard quadratic error loss with a quantile loss term to explicitly capture rare events while preserving the theoretical grounding of the original denoising objective.  We extensively evaluated our approach on a real-world credit card transaction dataset characterized by extreme class imbalance. The results demonstrate that the integration of diffusion-based synthetic data generation with a quantile-regularized denoising objective provides a robust and effective framework for fraud detection in highly imbalanced datasets.
Our code is available \href{https://github.com/abutalha-maths/quantile-tabddpm}{here}.

\keywords{Class Imbalanced Problems \and Diffusion Models \and  Quantile Loss \and Credit Card Fraud Detection.}
\end{abstract}
\section{Introduction}\label{sec1}
The focus of this paper is the \textit{credit card fraud detection} (CCFD) problem. The problem is a canonical and high-impact example of extreme class imbalance in real-world machine learning systems. In class imbalance problems, the minority class contains far fewer examples than the majority class. This problem is widespread in various domains, including textual tasks (e.g. spam detection, toxic or hate speech classification), medical imaging, and the present interest of the paper, that is, CCFD \cite{ibrahim2018imbalanced,makki2019experimental,mosquera2024class}. If the minority class is not distinguished correctly from the majority class, it can have severe real-world consequences. In many critical applications, such as healthcare, the minority class typically represents rare but high-risk conditions. For example, failing to correctly identify a patient with a disease may result in delayed or missed treatment, potentially leading to disease progression, complications, or even life-threatening outcomes \cite{araf2024cost}. Therefore, ensuring accurate detection of minority class instances is essential for building reliable and trustworthy predictive systems.

The rapid growth of financial services and e-commerce has made credit card fraud a major security concern \cite{hafez2025systematic}. Publicly available credit card fraud datasets remain limited because financial institutions restrict data sharing to protect user privacy. Moreover, fraudsters continuously evolve their techniques, making fraudulent transactions increasingly complex and difficult to detect \cite{lucas2020credit}. Balanced class distributions can be achieved through undersampling the majority class, oversampling the minority class, combining both approaches, or employing more advanced sampling techniques \cite{batista2004study,guo2008class}. 

Oversampling techniques, such as SMOTE, mitigate class imbalance by generating new samples that are interpolated between existing minority instances \cite{bahnsen2016feature,chawla2002smote}. Due to the highly skewed distribution, standard supervised learning algorithms tend to be biased towards the majority class, resulting in misleadingly high accuracy but poor detection of minority class \cite{he2009learning}. In the context of CCFD, misclassifying a fraudulent transaction as legitimate (i.e., a false negative) can result in direct financial losses to financial institutions or cardholders, making false negatives significantly more costly than false positives \cite{bahnsen2015example}.

Although these methods improve the representation of the minority class, they often fail to capture the complex and irregular data distributions found in real-world scenarios \cite{batista2004study,he2009learning}.

Conventional machine learning models, such as Logistic Regression, Decision Trees, Random Forests, and ensemble methods, have been widely studied for credit card fraud detection \cite{lebichot2020deep,randhawa2018credit,xuan2018random}. However, they often struggle with severe class imbalance and the constantly evolving nature of fraudulent behaviors. Common data-level approaches, including undersampling and oversampling (e.g., SMOTE), attempt to balance the class distribution, but each has inherent limitations \cite{ileberi2021performance}. SMOTE has also been extended to imbalanced regression: WSMOTER~\cite{camacho2024wsmoter} selects seed instances by density-based rarity weights instead of a fixed rare or common threshold.

To address these challenges, researchers have turned to generative approaches such as Generative Adversarial Nets (GANs) and Variational Autoencoders (VAEs) to synthesize realistic fraudulent transactions by learning the underlying data distribution \cite{shen2021credit}. Such synthetic data augmentation improves minority-class representation and enhances classifier performance under severe class imbalance. However, GANs often suffer from unstable training and mode collapse, whereas VAEs tend to generate overly smooth samples that fail to capture the subtle characteristics of fraudulent transactions \cite{goodfellow2014generative,kingma2014auto}. These limitations motivate the need for more robust generative approaches.

In recent years, \textit{Denoising Diffusion Probabilistic Models} (DDPMs) have attracted considerable attention for generative modeling \cite{ho2020denoising,sohl2015deep}. Their ability to capture complex data distributions makes them a promising alternative for synthetic data generation in imbalanced learning \cite{dhariwal2021diffusion,mukhopadhyay2023diffusion}. While DDPMs were originally developed for continuous, high-dimensional image data \cite{ho2020denoising}, tabular data present additional challenges due to their heterogeneous feature types, including continuous, categorical, numerical, and binary variables \cite{li2025diffusion,mueller2025continuous}.

Generative models, VAEs, GANs, and diffusion models, aim to learn the underlying data distribution to generate realistic synthetic samples. VAEs often produce overly smooth samples, while GANs are prone to training instability and mode collapse \cite{goodfellow2014generative,kingma2014auto}. Diffusion models overcome many of these limitations by learning a stable, likelihood-based denoising process that gradually transforms noise into data, resulting in improved sample quality and mode coverage \cite{ho2020denoising,sohl2015deep}.

Recently, several diffusion models have been proposed for tabular data \cite{kotelnikov2023tabddpm,liu2024controllable}. Among them, \textit{TabDDPM: Modelling Tabular Data with Diffusion Models} has emerged as a prominent approach by providing a principled framework for modeling mixed-type tabular data through an appropriate noise schedule and feature-specific transformations \cite{kotelnikov2023tabddpm}. Unlike standard DDPMs, TabDDPM explicitly models heterogeneous tabular features, including continuous, categorical, and binary variables, making it well suited for real-world mixed-type datasets \cite{ho2020denoising,kotelnikov2023tabddpm}.

Despite its many advantages, TabDDPM was not specifically designed for highly imbalanced learning and may struggle to accurately model minority fraud patterns in datasets with extreme class imbalance \cite{dal2015calibrating}. Its denoising objective is based on the standard \textit{quadratic error loss}, which assumes symmetric Gaussian noise and penalizes large deviations quadratically. As a result, the learned denoising process can be sensitive to outliers, particularly in credit card fraud data, where fraudulent transactions often reside in heavy-tailed, non-Gaussian regions of the feature space \cite{koenker2005quantile}. These limitations motivate the incorporation of quantile regularization into the denoising objective to improve robustness while preserving the underlying diffusion framework. The complete flow of the proposed Quantile-TabDDPM framework is illustrated in Figure~\ref{fig:flow-diagram} in the Appendix~\ref{app:flow}. 

\noindent{\textsc{Contribution of the paper.}}
Standard diffusion models, including TabDDPM, use a mean squared error (MSE) denoising objective, which can be sensitive to outliers and may underrepresent minority fraud patterns in highly imbalanced datasets. To address this limitation, we propose \textit{Quantile-TabDDPM}, which incorporates \textit{quantile regularization} into the TabDDPM denoising objective while preserving the original diffusion framework. Quantile loss, originating from quantile regression, is known for its robustness to heavy-tailed noise and extreme observations, making it well suited for modeling rare fraudulent transactions \cite{gu2016high,koenker2005quantile}. To the best of our knowledge, this is the first work to incorporate quantile regularization into a diffusion-based tabular generative model for highly imbalanced credit card fraud detection. Extensive experiments demonstrate that Quantile-TabDDPM consistently outperforms existing data augmentation approaches for credit card fraud detection. 

\section{Background Study}\label{sec2}

\subsection{TabDDPM for Tabular Data Generation}

TabDDPM extends the DDPM framework to tabular data by addressing challenges arising from heterogeneous feature types, multimodal feature distributions, and complex feature dependencies \cite{kotelnikov2023tabddpm}. It employs dedicated mechanisms for continuous and categorical variables while preserving the probabilistic diffusion framework. Since our proposed method builds upon TabDDPM, we briefly review the underlying DDPM formulation.

\paragraph{Gaussian Diffusion Models}

A DDPM consists of a forward diffusion process that gradually corrupts the input data with Gaussian noise and a reverse denoising process that learns to reconstruct clean samples from noisy observations \cite{ho2020denoising}.

Given an input sample $\mathbf{x}_0$, the forward process admits the following closed-form representation:

\begin{equation}
\mathbf{x}_t =
\sqrt{\bar{\alpha}_t}\mathbf{x}_0
+
\sqrt{1-\bar{\alpha}_t}\boldsymbol{\epsilon},
\qquad
\boldsymbol{\epsilon}\sim\mathcal{N}(0,\mathbf{I}),
\label{eq:forward_closedform}
\end{equation}

where $\beta_t:=1-\alpha_t$ and
$\bar{\alpha}_t=\prod_{i=1}^{t}\alpha_i$ define the diffusion noise schedule.

The reverse process learns the conditional distribution

\begin{equation}
p_\theta(\mathbf{x}_{t-1}|\mathbf{x}_t)
=
\mathcal{N}
(\mathbf{x}_{t-1};
\mu_\theta(\mathbf{x}_t,t),
\sigma_t^2\mathbf{I}),
\label{eq:reverse_process}
\end{equation}

where the reverse-process mean is given by

\begin{equation}
\mu_\theta(\mathbf{x}_t,t)
=
\frac{1}{\sqrt{\alpha_t}}
\left(
\mathbf{x}_t
-
\frac{\beta_t}{\sqrt{1-\bar{\alpha}_t}}
\epsilon_\theta(\mathbf{x}_t,t)
\right).
\label{eq:reverse_mean}
\end{equation}

Here, $\epsilon_\theta$ denotes the neural network that predicts the Gaussian noise added during the forward diffusion process and enables iterative sample generation through the reverse denoising procedure. 
The training objective is the simplified denoising loss:

\begin{equation}
\label{eq:mse_simple}
L^{\text{simple}}_t
=
\mathbb{E}_{x_0,\epsilon,t}
\left\|
\epsilon-\epsilon_\theta(x_t,t)
\right\|_2^2.
\end{equation}

where $\epsilon_\theta$ is trained to predict the Gaussian noise added during the forward diffusion process. \\

TabDDPM extends DDPM to handle mixed-type tabular data by modeling both continuous and categorical features within a unified diffusion framework \cite{kotelnikov2023tabddpm}. Numerical features correspond to continuous-valued variables represented by real numbers, whereas categorical features take values from a finite set of discrete categories.

\begin{itemize}
    \item Numerical features are modeled using Gaussian diffusion with quantile normalization.
    \item Categorical features are represented using one-hot encoding and modeled through multinomial diffusion.
\end{itemize}

Gaussian diffusion is naturally suited for numerical features because it operates in continuous space by gradually corrupting real-valued variables with additive Gaussian noise \cite{ho2020denoising,kingma2021variational}. The corresponding forward and reverse diffusion processes are described in Equations~(\ref{eq:forward_closedform}--\ref{eq:mse_simple}).

For categorical features, TabDDPM employs multinomial diffusion, where each categorical variable is represented as a one-hot encoded vector and progressively corrupted through a categorical transition process \cite{hoogeboom2021argmax}. This enables the model to effectively capture discrete feature distributions while preserving the probabilistic diffusion framework.

Accordingly, each tabular sample is represented as
\[
x=\left[x_{\text{num}},x_{\text{cat}_1},\ldots,x_{\text{cat}_C}\right],
\]
where $x_{\text{num}}$ denotes the numerical features and $x_{\text{cat}_i}$ denotes the $i$-th categorical feature. The denoising network jointly predicts the Gaussian noise associated with numerical features and the categorical probabilities required for multinomial diffusion.

The original TabDDPM training objective combines the Gaussian denoising loss for numerical features with multinomial diffusion objectives for categorical features \cite{kotelnikov2023tabddpm}. The reverse diffusion process is implemented using a multi-layer perceptron (MLP)-based denoising network with sinusoidal timestep embeddings and optional class-label embeddings for conditional generation \cite{gorishniy2021revisiting,dhariwal2021diffusion,nichol2021improved}. 
Since the proposed Quantile-TabDDPM builds directly upon this framework, our modification incorporates quantile regularization into the Gaussian denoising objective while preserving the original diffusion formulation and the overall TabDDPM architecture.

\subsection{Quantile Loss and Motivation}

Quantile regression extends classical least squares by estimating conditional quantiles rather than only the conditional mean \cite{koenker2001quantile}. Consequently, quantile loss (also known as the pinball loss) provides a robust alternative to quadratic loss, particularly for skewed, heavy-tailed, and heteroscedastic data distributions. For a quantile level $\tau\in(0,1)$, predicted value $\hat{y}$, and target value $y$, the quantile loss is defined as

\begin{equation}
\mathcal{Q}_{\tau}(y,\hat{y})
=
(\tau-\mathbb{I}_{\{y<\hat{y}\}})
(y-\hat{y}),
\label{eq:quantile_loss}
\end{equation}

where $\mathbb{I}_{\{\cdot\}}$ denotes the indicator function.

Unlike quadratic loss, quantile loss is piecewise linear and reduces the influence of large errors, making it robust to outliers, heavy-tailed noise, and heteroscedasticity \cite{koenker2005quantile,koenker2001quantile}. Since fraudulent transactions frequently occur in the tails of the underlying data distribution, quantile-based learning provides a principled mechanism for emphasizing these rare but informative observations \cite{dal2015calibrating,koenker2001quantile}. This motivates the incorporation of \emph{quantile regularization} into the TabDDPM denoising objective while preserving the original diffusion formulation.

\section{Methodology}

\subsection{Problem Setup}

We consider a labeled dataset $D=\{(x_i,y_i)\}_{i=1}^{N}$, where $x_i\in\mathbb{R}^{d}$ denotes the feature vector of the $i$-th sample and $y_i\in\{0,1\}$ is the corresponding class label, with $y_i=1$ representing the minority (fraud) class and $y_i=0$ the majority (non-fraud) class.

\subsection{A New Proposal: Quantile-TabDDPM}

The proposed Quantile-TabDDPM builds directly upon the TabDDPM framework by incorporating \emph{quantile regularization} into the Gaussian denoising objective while preserving the original diffusion formulation and overall model architecture \cite{kotelnikov2023tabddpm}.

TabDDPM provides a diffusion-based framework for generative modeling of tabular data but was originally developed for general tabular learning rather than highly imbalanced classification. In its original formulation, numerical features are optimized using the quadratic denoising objective in Equation~(\ref{eq:mse_simple}), which primarily targets the conditional mean. This objective can be less effective for highly skewed and heavy-tailed data, where rare minority-class observations contribute only weakly during training. To improve robustness, we incorporate the quantile loss defined in Equation~(\ref{eq:quantile_loss}) as a regularization term within the Gaussian denoising objective, while leaving the remaining diffusion framework unchanged. Throughout this work, we consider datasets containing only numerical features, which are modeled as continuous real-valued variables.

The quantile-regularized denoising objective of Quantile-TabDDPM is defined as

\begin{equation}
\mathcal{L}_t^{\text{Quantile-TabDDPM}}
=
\mathbb{E}_{x_0,\epsilon,t}
\left[
\|\epsilon-\epsilon_\theta(x_t,t)\|_2^2
+
\lambda\,
\mathcal{Q}_{\tau}
\bigl(\epsilon,\epsilon_\theta(x_t,t)\bigr)
\right],
\label{eq:quantile_tabddpm_loss}
\end{equation}

where $\mathcal{Q}_{\tau}(\cdot)$ denotes the quantile loss defined in Equation~(\ref{eq:quantile_loss}), and $\lambda$ controls the contribution of the quantile regularization term. The value of $\lambda$ is selected through hyperparameter tuning.

\begin{figure}[H]
\centering
\begin{subfigure}[b]{0.49\textwidth}
    \centering
    \includegraphics[height=2.1cm,width=\linewidth,keepaspectratio]{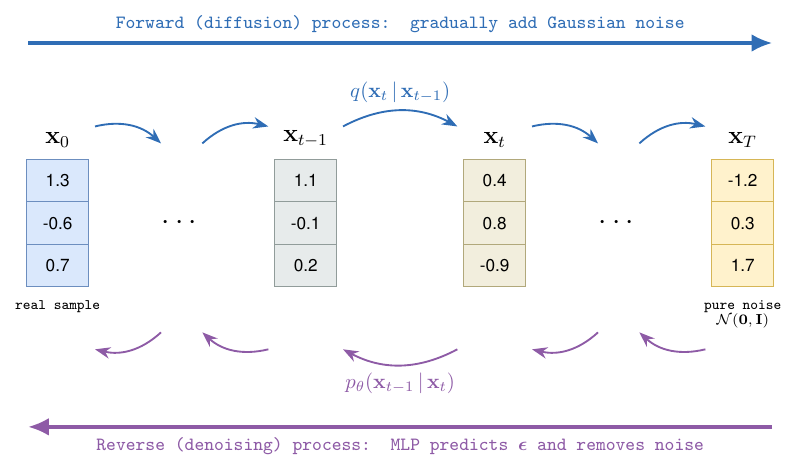}
    \caption{Forward and reverse processes in DDPM.}
    \label{fig:ddpm_fwd_rev}
\end{subfigure}\hfill
\begin{subfigure}[b]{0.49\textwidth}
    \centering
    \includegraphics[height=3.5cm,width=\linewidth,keepaspectratio]{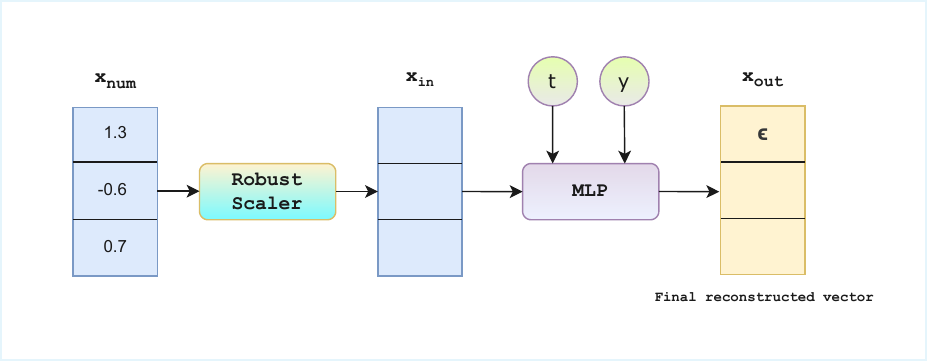}
    \caption{Denoising network of Quantile-TabDDPM.}
    \label{fig:qtabddpm_reverse}
\end{subfigure}
\caption{Overview of the proposed Quantile-TabDDPM framework for class-conditional generation. Here, $t$ denotes the diffusion timestep and $y$ represents the class label used for conditional generation following the TabDDPM architecture.}
\label{fig:ddpm_overview}
\end{figure}

The quadratic loss preserves the original denoising objective of TabDDPM, while the quantile regularization term improves robustness to skewed, heavy-tailed data by placing greater emphasis on rare but informative prediction errors. An overview of the proposed Quantile-TabDDPM framework is shown in Fig.~\ref{fig:ddpm_overview}.

\section{Experimental Setup and Results}\label{sec4}

The proposed Quantile-TabDDPM is evaluated on the European Credit Card (ECC) dataset using standard evaluation metrics for imbalanced classification \cite{ecc_dataset}. The corresponding experimental results are presented in this section.

\subsection{Dataset Description}

We evaluate the proposed Quantile-TabDDPM on the ECC dataset \cite{ecc_dataset}, which contains transaction records collected from European cardholders in September 2013. The dataset consists exclusively of numerical features and contains 284,807 transactions, of which only 492 are fraudulent (0.172\%), making it a standard benchmark for highly imbalanced credit card fraud detection.

\subsection{Evaluation Metrics}

Model performance is evaluated using Precision, Recall, F1-score, ROC-AUC, and Matthews Correlation Coefficient (MCC), which are widely adopted for imbalanced classification \cite{chicco2020advantages,he2009learning}. Since accuracy can be misleading under severe class imbalance, these metrics provide a more reliable assessment of fraud detection performance by jointly evaluating minority-class detection and false alarms.

\subsection{Data Preprocessing and Model Input Pipeline}

The ECC dataset contains 30 numerical features and a binary class label \cite{ecc_dataset}. Since all input features are numerical, each feature is standardized using the \texttt{RobustScaler} from \texttt{scikit-learn}, which scales data based on the median and interquartile range to improve robustness against outliers and heavy-tailed feature distributions \cite{pedregosa2011scikit}. The transformed features are subsequently clipped to the range $[-10,10]$ to improve numerical stability during diffusion training.

For class-conditional generation, the binary class labels are encoded as one-hot vectors and concatenated with the standardized feature vector before being provided to the MLP-based diffusion network (Fig.~\ref{fig:ddpm_overview}). The proposed Quantile-TabDDPM is designed for numerical tabular data; therefore, we evaluate it on the ECC dataset, which contains exclusively continuous numerical features.

\subsection{A New Approach: Data Augmentation}

The proposed Quantile-TabDDPM is trained using the 80\%/10\%/10\% stratified train-validation-test split. The diffusion model is trained only on the training set, and synthetic samples are generated exclusively for the minority (fraud) class. These samples are then combined with the original training data to partially alleviate class imbalance while leaving all legitimate transactions unchanged.

Rather than enforcing a fully balanced dataset, we adopt a 90:10 non-fraud-to-fraud ratio for data augmentation. This strategy improves minority-class representation while preserving a realistic class distribution \cite{dal2015calibrating,weston2008plastic}. We empirically evaluated multiple augmentation ratios and found that a 90:10 non-fraud-to-fraud ratio provided the best trade-off between minority-class representation and classification performance. Consequently, this ratio is used throughout the experiments.

\begin{figure*}[!t]
    \centering

    \begin{subfigure}[b]{0.32\textwidth}
        \centering
        \caption{V12}
        \includegraphics[width=\linewidth]{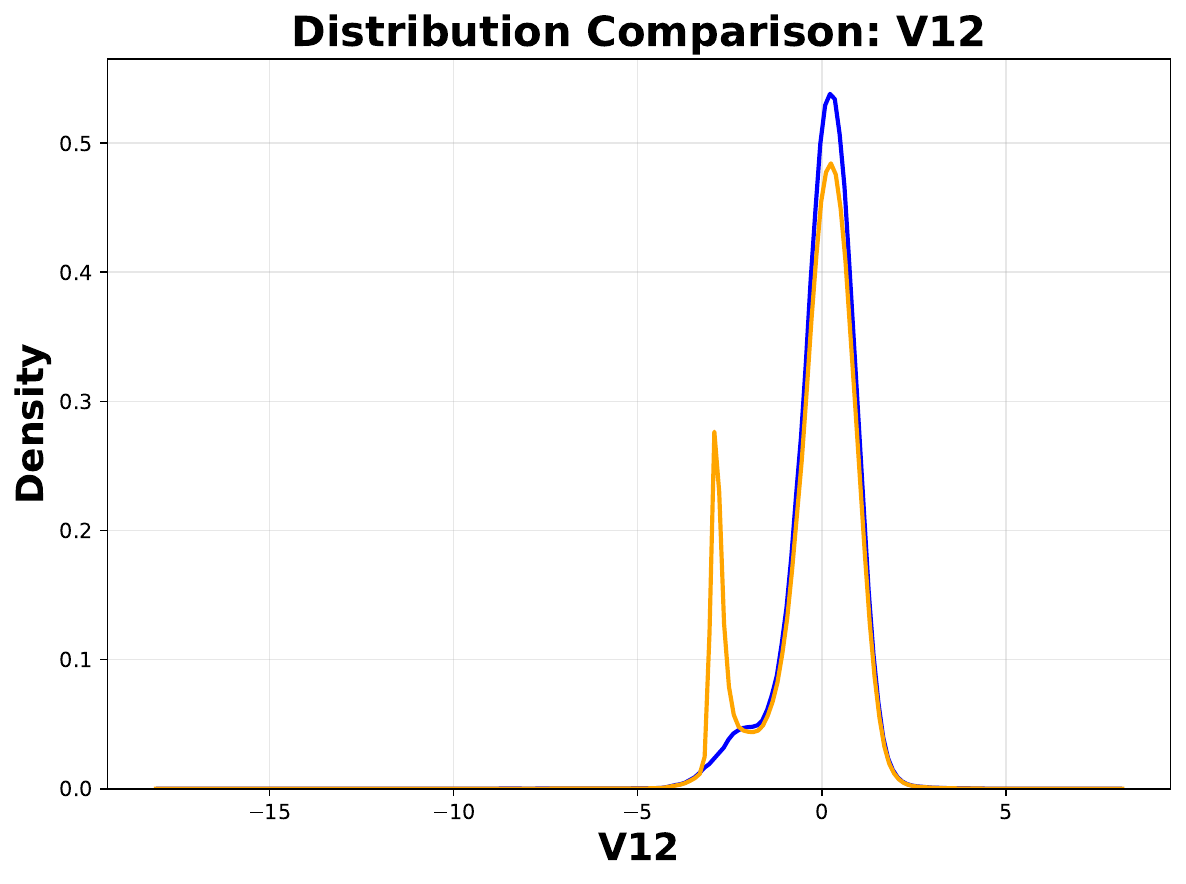}
        
        \label{fig:q_feat2}
    \end{subfigure}
    \hfill
    \begin{subfigure}[b]{0.32\textwidth}
        \centering
        \caption{V16}
        \includegraphics[width=\linewidth]{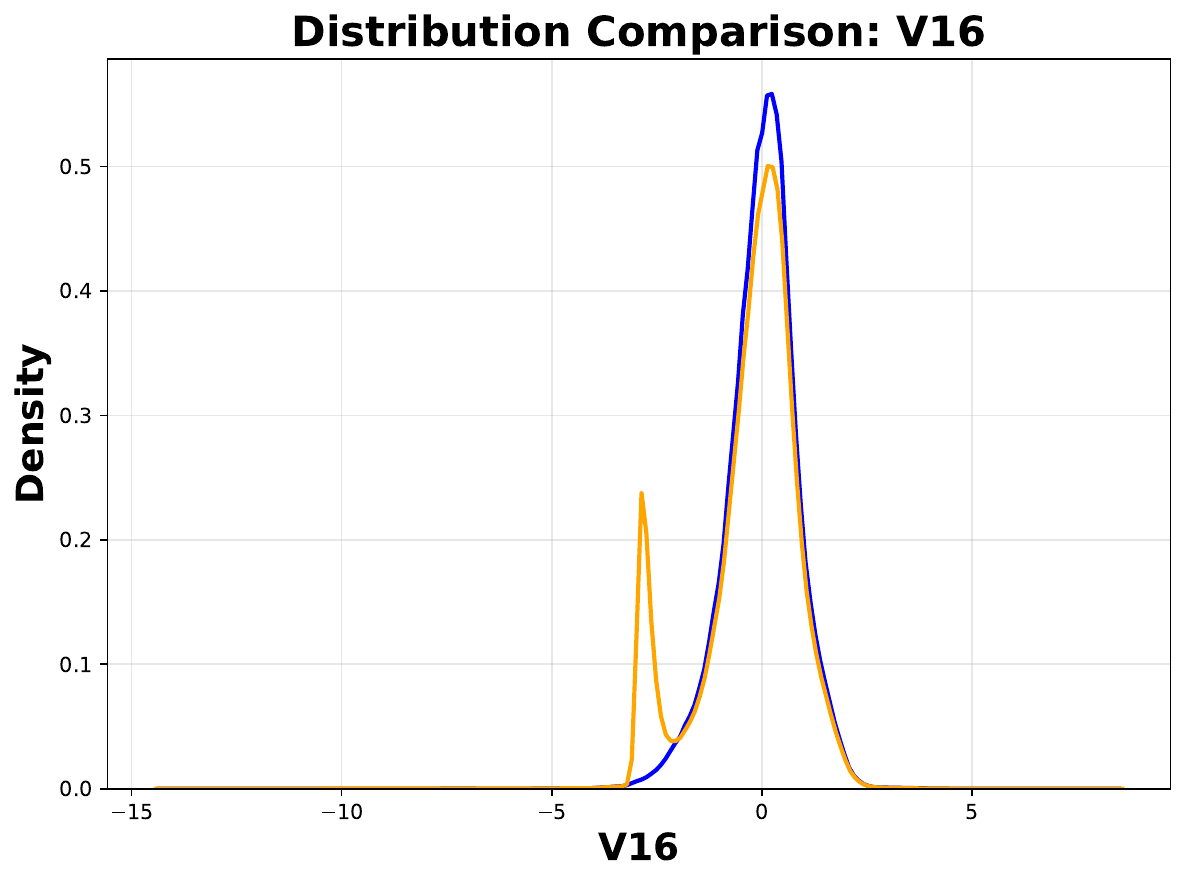}
        
        \label{fig:q_feat3}
    \end{subfigure}
    \hfill
    \begin{subfigure}[b]{0.32\textwidth}
        \centering
        \caption{V18}
        \includegraphics[width=\linewidth]{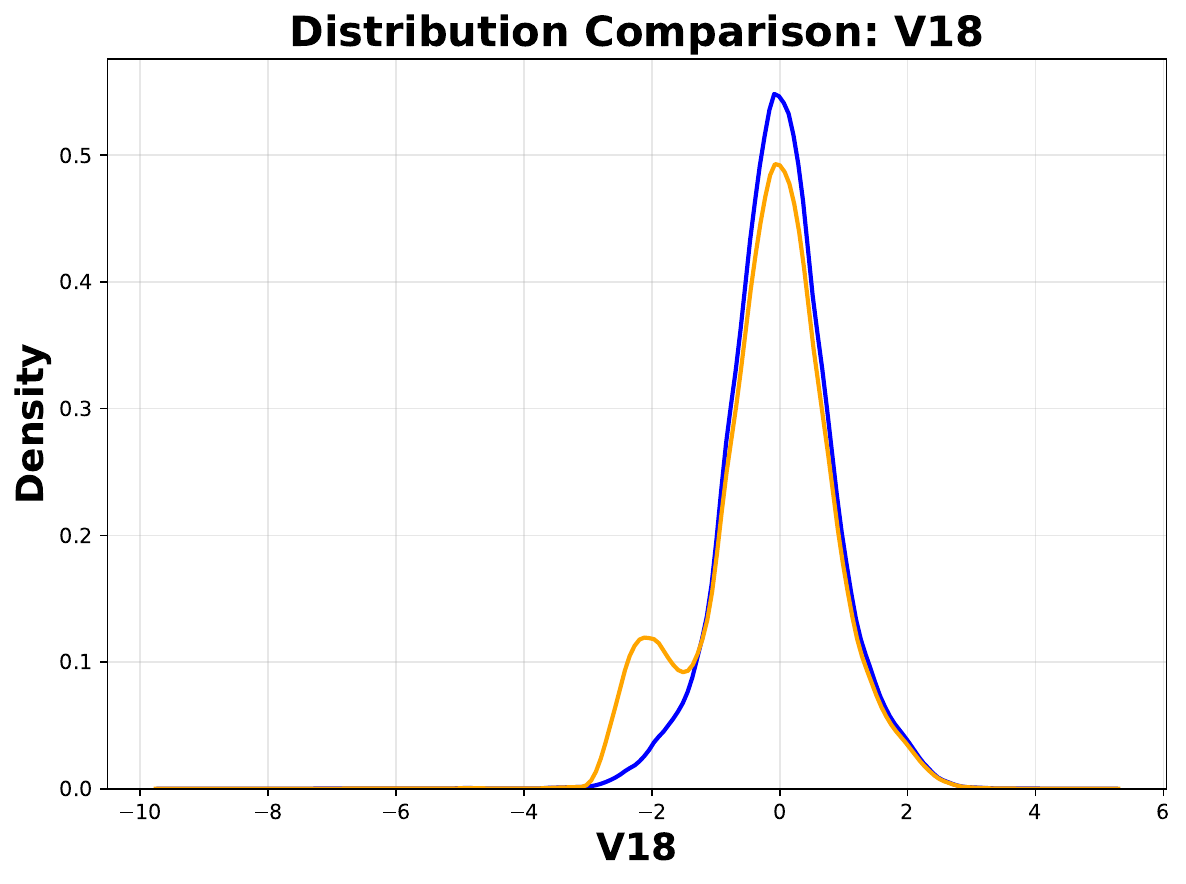}
        
        \label{fig:q_feat4}
    \end{subfigure}

    \vspace{0.8em}

    \scriptsize{(a) Quantile-TabDDPM}

    \vspace{1.2em}

    \begin{subfigure}[b]{0.32\textwidth}
        \centering
        \caption{V12}
        \includegraphics[width=\linewidth]{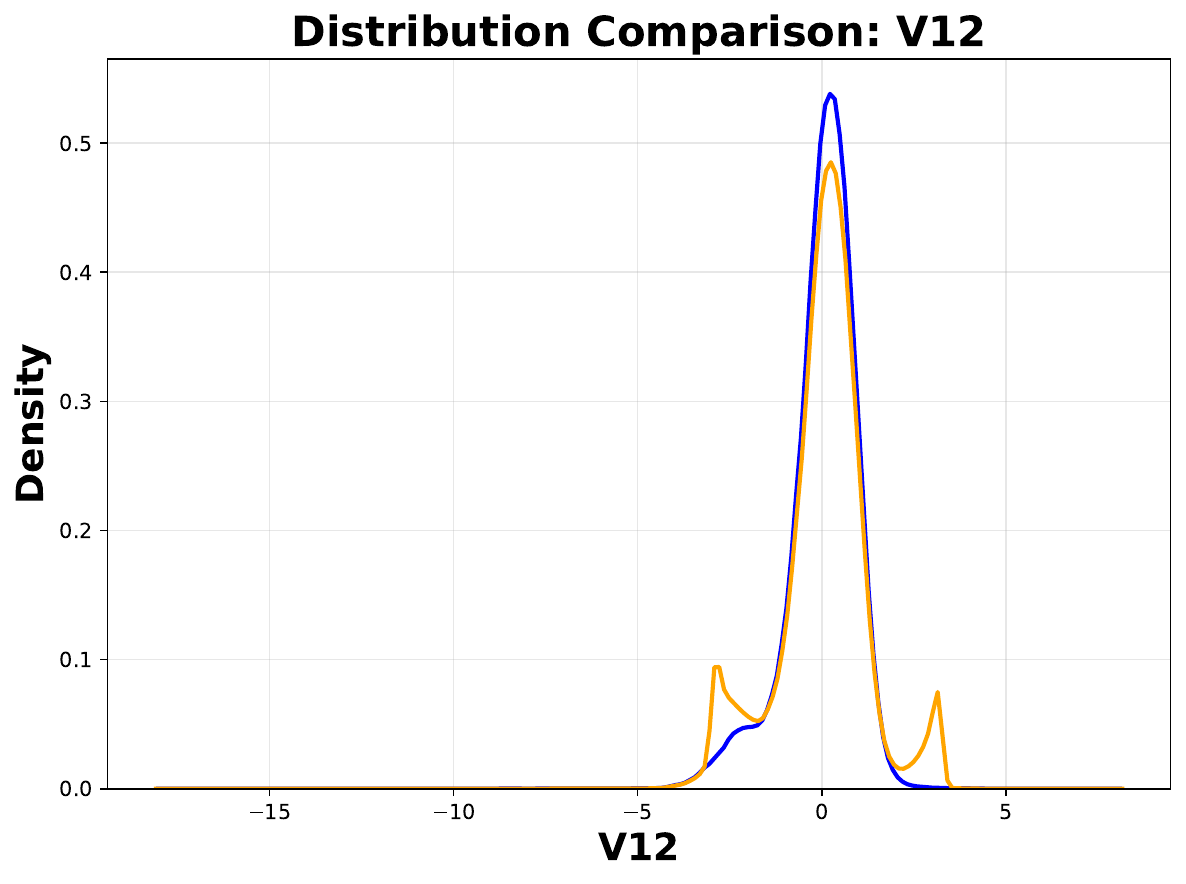}
        
        \label{fig:b_feat2}
    \end{subfigure}
    \hfill
    \begin{subfigure}[b]{0.32\textwidth}
        \centering
        \caption{V16}
        \includegraphics[width=\linewidth]{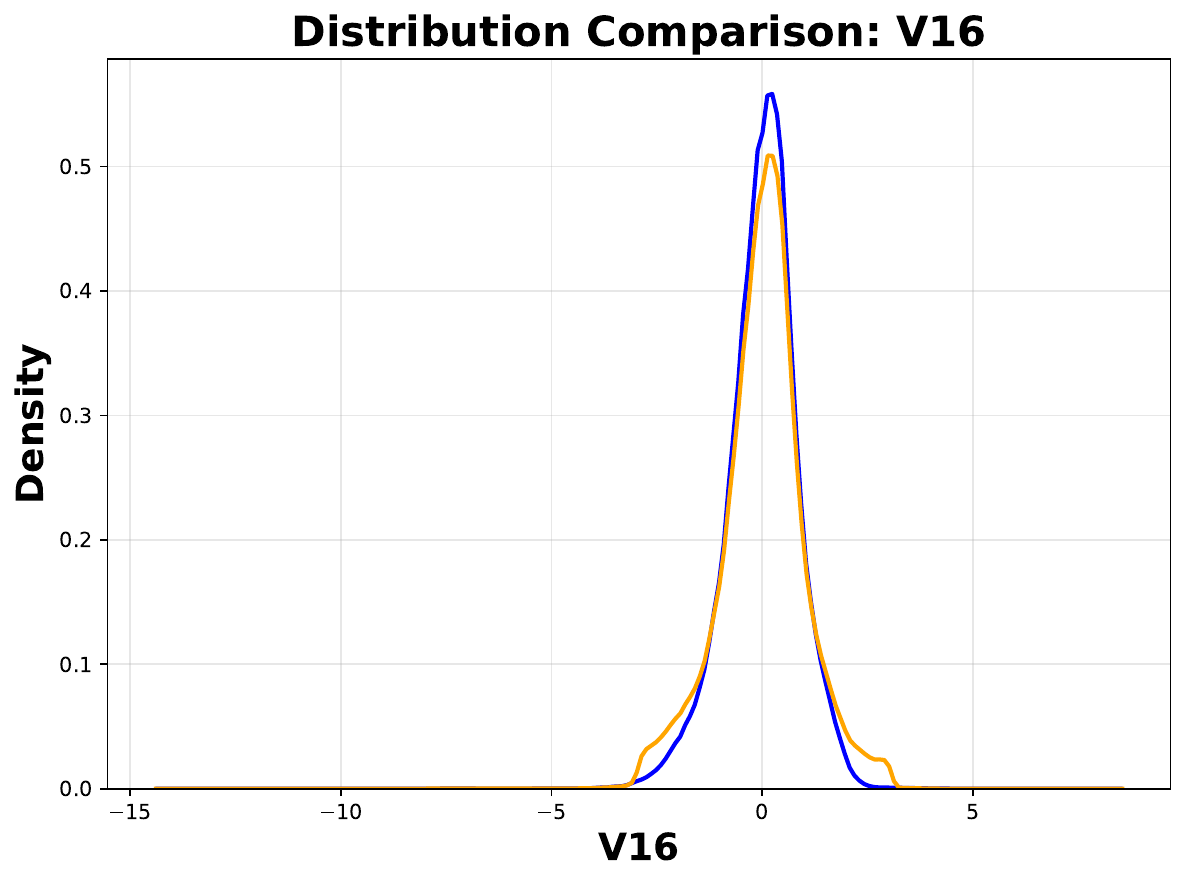}
        
        \label{fig:b_feat3}
    \end{subfigure}
    \hfill
    \begin{subfigure}[b]{0.32\textwidth}
        \centering
        \caption{V18}
        \includegraphics[width=\linewidth]{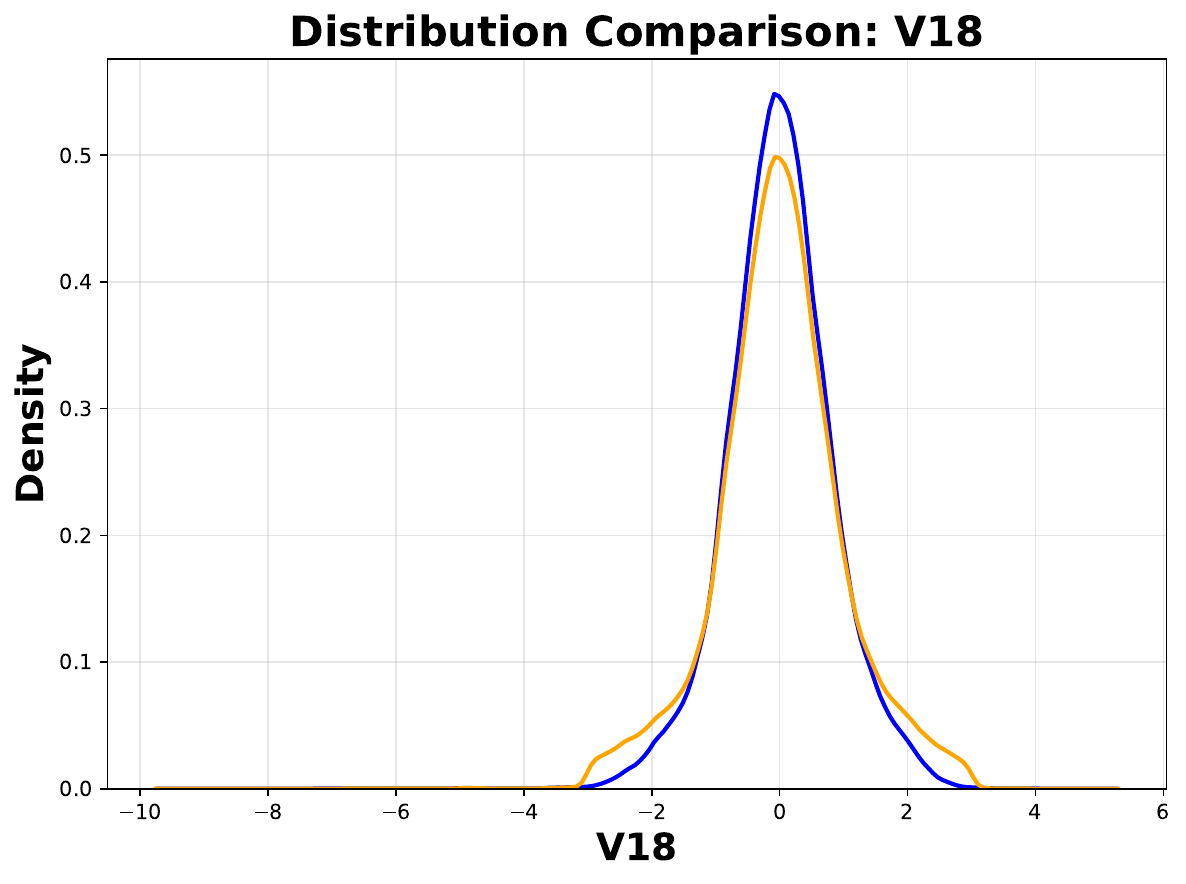}
        
        \label{fig:b_feat4}
    \end{subfigure}

    \vspace{0.8em}

    \scriptsize{(b) TabDDPM (Baseline Model)}
    \vspace{0.3cm}
\caption{Feature-wise distribution comparison between original training data and augmented data (90:10 data augmentation strategy). The blue curves correspond to the original training data, while the yellow curves represent augmented samples.
    In the first row, augmentation is generated using Quantile-TabDDPM, whereas in the second row, it is generated using the baseline TabDDPM.
    Each column represents the same feature, enabling direct visual comparison of distributional fidelity across models.}
    
    \label{fig:featurewise_dist_comparison}
\end{figure*}

To assess the quality of the generated samples, Fig.~\ref{fig:featurewise_dist_comparison} compares the feature distributions of the original training data and the augmented dataset. Quantile-TabDDPM preserves the overall feature distributions while introducing additional minority-class samples, as evidenced by the strong agreement between the original and augmented distributions. In contrast, the baseline TabDDPM exhibits less faithful distribution matching after augmentation. See Appendix \ref{app:results} for additional visualization results.

The training set was used to train Quantile-TabDDPM and generate synthetic fraud samples, while the validation set was used for hyperparameter tuning. Tree-based classifiers were then trained on the augmented training data and evaluated on the held-out test set. We employ tree-based classifiers due to their strong performance on imbalanced tabular data \cite{breiman2001random,chen2016xgboost,he2009learning}.

\subsection{Hyperparameter Tuning Process}

The hyperparameters of Quantile-TabDDPM were optimized using Optuna \cite{akiba2019optuna}. The optimization objective was the F1-score of an XGBoost classifier trained on the augmented training set and evaluated on the validation set. We jointly optimized the learning rate, batch size, diffusion timesteps, MLP architecture, dropout rate, quantile level $\tau$, quantile weight $\lambda$, synthetic sample proportion, and noise schedule. The hyperparameter search space and the selected values are summarized in Table~\ref{tab:hyperparameters}.

\begin{table}[H]
\centering
\footnotesize
\setlength{\tabcolsep}{4pt}
\renewcommand{\arraystretch}{0.95}
\caption{Hyperparameter search space and optimized values for Quantile-TabDDPM on the ECC dataset.}
\label{tab:hyperparameters}
\begin{tabular}{lll}
\toprule
\textbf{Hyperparameter} & \textbf{Search Space} & \textbf{Selected} \\
\midrule
Learning rate         & Log-uniform $[10^{-5},\,3\times10^{-3}]$ & 0.00014106 \\
Batch size            & $\{256,512,1024,4096\}$ & 512 \\
Diffusion timesteps   & $\{100,1000\}$ & 1000 \\
Training epochs       & Fixed (1500) & 1500 \\
Number of MLP layers  & Integer $[2,8]$ (step = 2) & 4 \\
Optimizer             & AdamW & AdamW \\
Scheduler             & $\{$cosine, linear, sigmoid$\}$ & Sigmoid \\
Proportion of samples & $\{0.25,0.5,1.0,2.0\}$ & 1.0 \\
Dropout               & 0.1 & 0.1 \\
Loss function         & Quantile loss & Quantile loss \\
Quantile level $\tau$ & $[0.6,0.9]$ (step = 0.05) & 0.75 \\
Quantile weight $\lambda$ & $[0.1, 1.0]$ & 0.2384 \\
Number of trials      & 50 & 50 \\
\bottomrule
\end{tabular}
\end{table}
For each Optuna trial, an XGBoost classifier was trained on the augmented training set and evaluated on the validation set using the F1-score as the optimization objective. After optimization, the final Quantile-TabDDPM model was trained using the selected hyperparameters summarized in Table~\ref{tab:hyperparameters}. The same configuration was used for all subsequent experiments to ensure a fair and consistent evaluation. Since the optimal hyperparameters depend on the characteristics of the underlying dataset, the reported configuration is specific to the ECC dataset and may require re-tuning for other application domains.

\subsection{Results and Analysis}\label{sec:analysis}

The proposed Quantile-TabDDPM is evaluated using a 90:10 non-fraud-to-fraud augmentation strategy on the held-out test set. We compare our approach against TabDDPM, SMOTE, CTGAN, and TVAE using multiple tree-based classifiers.

\begin{figure}[!t]
    \centering

    \begin{subfigure}[t]{0.49\textwidth}
        \centering
        \includegraphics[width=\linewidth]{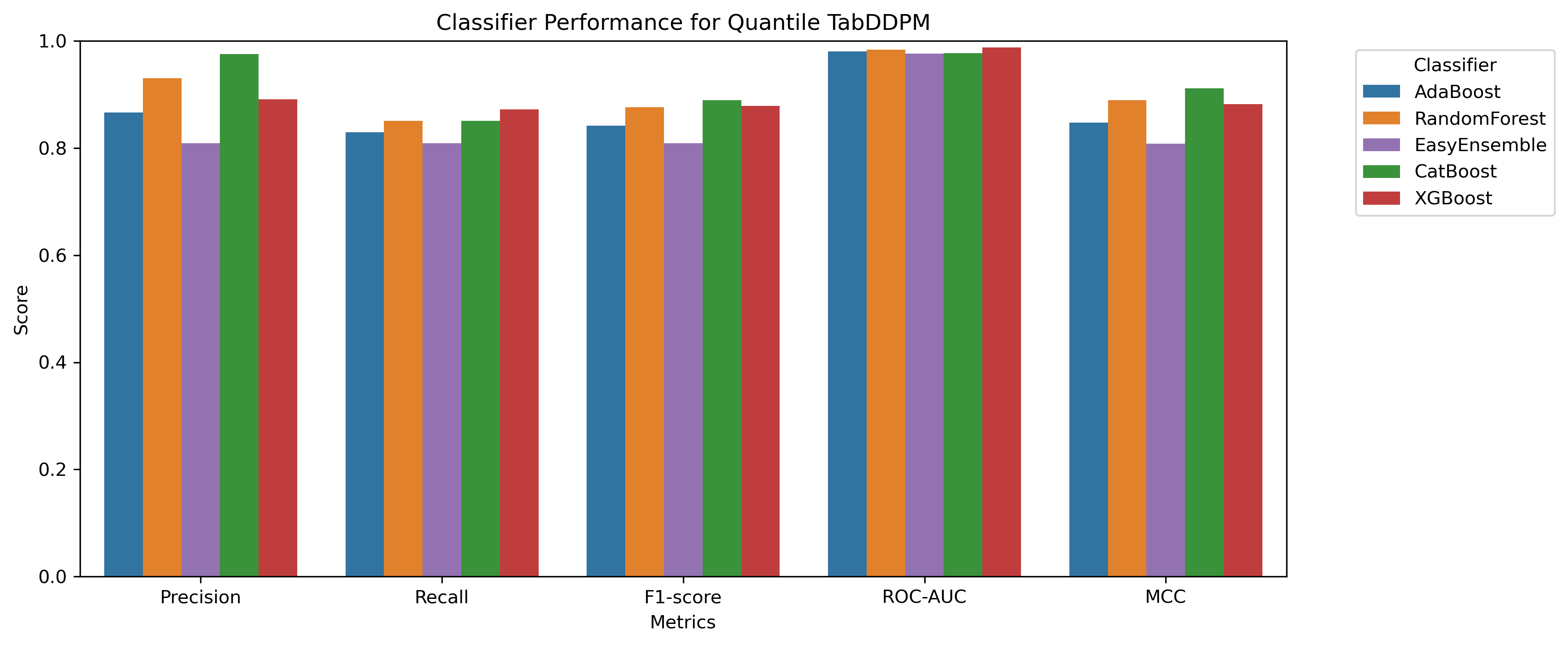}
        \caption{Quantile-TabDDPM (Ours)}
        \label{fig:metrics_quantile}
    \end{subfigure}
    \hfill
    \begin{subfigure}[t]{0.49\textwidth}
        \centering
        \includegraphics[width=\linewidth]{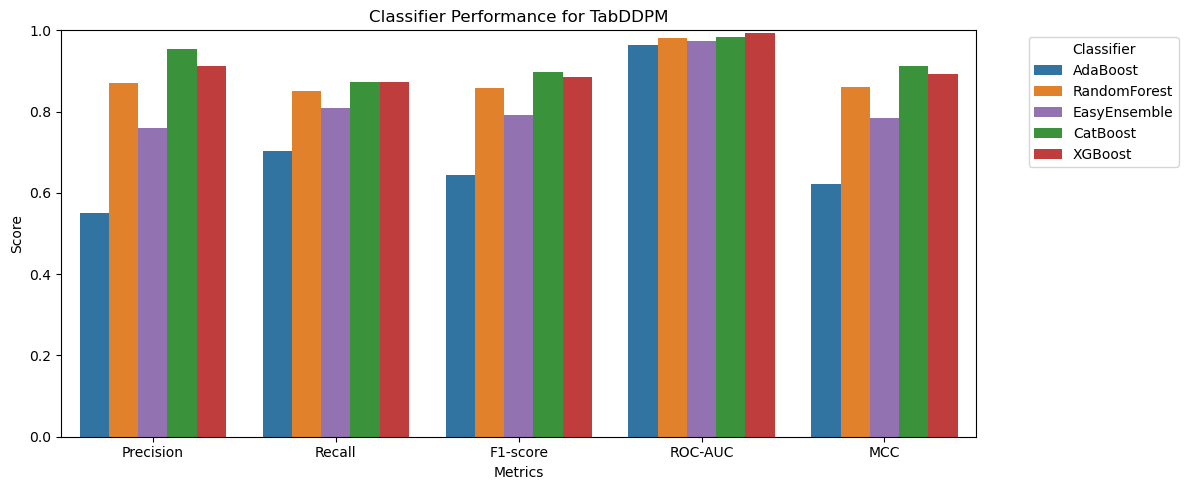}
        \caption{TabDDPM (Baseline)}
        \label{fig:metrics_tabddpm}
    \end{subfigure}

    \caption{Performance comparison of tree-based classifiers on the ECC dataset using the 90:10 non-fraud-to-fraud augmentation strategy. The bar charts summarize Precision, Recall, F1-score, ROC-AUC, and MCC for (a) Quantile-TabDDPM and (b) the baseline TabDDPM, evaluated on the held-out test set.}
    \label{fig:classifier_comparison}
\end{figure}

 Fig.~\ref{fig:classifier_comparison} visually compare the performance of the evaluated classifiers, further demonstrating the consistent improvements achieved by Quantile-TabDDPM over the baseline TabDDPM across multiple evaluation metrics.

\begin{table}[!t]
\centering
\scriptsize
\setlength{\tabcolsep}{3pt}
\caption{Performance comparison of Quantile-TabDDPM, TabDDPM (90:10 non-fraud-to-fraud ratio), and SMOTE.}
\label{tab:quantile_tabddpm}
\label{tab:smote_performance}
\resizebox{\columnwidth}{!}{%
\begin{tabular}{lccccccccccccccc}
\toprule
\multirow{2}{*}{Classifier} &
\multicolumn{5}{c}{Quantile-TabDDPM (Ours)} &
\multicolumn{5}{c}{TabDDPM} &
\multicolumn{5}{c}{SMOTE} \\
\cmidrule(lr){2-6}\cmidrule(lr){7-11}\cmidrule(lr){12-16}
& Prec. & Rec. & F1 & AUC & MCC &
Prec. & Rec. & F1 & AUC & MCC &
Prec. & Rec. & F1 & AUC & MCC \\
\midrule
AdaBoost     & \textbf{0.867} & \textbf{0.829} & \textbf{0.842} & \textbf{0.988} & \textbf{0.848} & 0.704 & 0.809 & 0.770 & 0.976 & 0.754 & 0.577 & 0.837 & 0.683 & 0.918 & 0.695 \\
RandomForest & \textbf{0.930} & \textbf{0.851} & \textbf{0.876} & 0.984 & \textbf{0.889} & 0.826 & 0.809 & 0.814 & \textbf{0.992} & 0.817 & \textbf{0.932} & 0.837 & \textbf{0.882} & 0.918 & \textbf{0.883} \\
EasyEnsemble & \textbf{0.809} & 0.809 & \textbf{0.809} & 0.976 & \textbf{0.808} & 0.500 & \textbf{0.851} & 0.690 & \textbf{0.990} & 0.652 & 0.055 & \textbf{0.939} & 0.104 & \textbf{0.956} & 0.224 \\
CatBoost     & \textbf{0.976} & 0.851 & \textbf{0.889} & 0.977 & \textbf{0.911} & 0.800 & 0.851 & 0.833 & \textbf{0.991} & 0.825 & 0.891 & 0.837 & 0.863 & 0.918 & 0.863 \\
XGBoost      & \textbf{0.891} & \textbf{0.872} & \textbf{0.879} & 0.988 & \textbf{0.882} & 0.886 & 0.830 & 0.848 & \textbf{0.990} & 0.857 & 0.911 & 0.837 & 0.872 & 0.918 & 0.873 \\
\bottomrule
\end{tabular}}
\end{table}

\begin{table}[!t]
\centering
\scriptsize
\setlength{\tabcolsep}{3pt}
\caption{Performance comparison of CTGAN and TVAE (90:10 non-fraud-to-fraud ratio).}
\label{tab:ctgan_tvae}
\resizebox{\columnwidth}{!}{%
\begin{tabular}{lcccccccccc}
\toprule
\multirow{2}{*}{Classifier} &
\multicolumn{5}{c}{CTGAN} &
\multicolumn{5}{c}{TVAE} \\
\cmidrule(lr){2-6}\cmidrule(lr){7-11}
& Prec. & Rec. & F1 & AUC & MCC &
Prec. & Rec. & F1 & AUC & MCC \\
\midrule
AdaBoost     & 0.284 & 0.891 & 0.431 & 0.982 & 0.503 & 0.408 & 0.867 & 0.556 & \textbf{0.985} & 0.595 \\
RandomForest & 0.816 & 0.869 & 0.842 & 0.982 & 0.842 & \textbf{0.854} & \textbf{0.891} & \textbf{0.872} & 0.970 & \textbf{0.872} \\
EasyEnsemble & 0.053 & \textbf{0.935} & 0.096 & \textbf{0.987} & 0.218 & 0.078 & \textbf{0.891} & 0.143 & 0.979 & 0.261 \\
CatBoost     & \textbf{0.837} & 0.891 & \textbf{0.863} & 0.978 & \textbf{0.863} & 0.796 & 0.848 & 0.821 & 0.984 & 0.821 \\
XGBoost      & 0.600 & 0.913 & 0.724 & 0.979 & 0.739 & 0.667 & 0.869 & 0.755 & 0.969 & 0.761 \\
\bottomrule
\end{tabular}}
\end{table}


Tables~\ref{tab:quantile_tabddpm} and \ref{tab:ctgan_tvae} show that Quantile-TabDDPM consistently achieves competitive and well-balanced performance across all classifiers. For example, with CatBoost, our method achieves the highest precision (0.976) and MCC (0.911), together with a strong F1-score (0.889), outperforming the baseline TabDDPM. Similarly, XGBoost attains an F1-score of 0.879 and an MCC of 0.882, exceeding the corresponding results obtained using CTGAN and TVAE.

In contrast, CTGAN and TVAE frequently exhibit unstable precision--recall trade-offs. For example, EasyEnsemble combined with CTGAN achieves a high recall (0.935) but an extremely low precision (0.053), indicating a large number of false positives. Although TabDDPM performs competitively, Quantile-TabDDPM generally achieves higher F1-scores and MCC values while maintaining comparable ROC-AUC performance.

Compared with SMOTE (Table~\ref{tab:smote_performance}), the proposed method provides a more balanced trade-off between precision and recall across classifiers. Overall, these results demonstrate that Quantile-TabDDPM generates high-quality synthetic fraud samples that consistently improve downstream classifier performance under severe class imbalance.

\subsection{Monte Carlo Simulations}

Monte Carlo simulations are used to assess the robustness and statistical reliability of experimental results by repeatedly running the same procedure under different random realizations \cite{robert2004monte}. Rather than relying on a single random train-test partition, this approach evaluates the degree to which the model performance is sensitive to variations in data splitting and stochastic training effects \cite{hastie2009elements,bishop2006prml}. 

We ran 10 independent Monte Carlo simulations to assess the consistency and reliability of our results. We combined the results and reported the average performance with the standard deviation in Table~\ref{tab:monte_carlo_results_10}. XGBoost leads, followed by RandomForest and CatBoost, matching our single-split results and confirming the Quantile-TabDDPM framework's consistency across random splits. Results from an extended analysis with 30 simulations, which show the same trend, are provided in Appendix~\ref{app:experimental}.

\begin{table}[htbp]
\centering
\caption{Monte Carlo simulation aggregate results across 10 independent runs for Quantile-TabDDPM with a 90:10 non-fraud-to-fraud ratio.}
\scriptsize 
\setlength{\tabcolsep}{3pt} 
\resizebox{\textwidth}{!}{%
\begin{tabular}{lccccc}
\toprule
Classifier & Precision & Recall & F1-score & ROC-AUC & MCC \\
\midrule
AdaBoost      & $0.831 \pm 0.040$ & $0.817 \pm 0.044$ & $0.820 \pm 0.024$ & $\mathbf{0.976 \pm 0.016}$ & $0.823 \pm 0.018$ \\
RandomForest  & $\mathbf{0.899 \pm 0.053}$ & $\mathbf{0.857 \pm 0.039}$ & $\mathbf{0.870 \pm 0.025}$ & $\mathbf{0.983 \pm 0.009}$ & $\mathbf{0.877 \pm 0.025}$ \\
EasyEnsemble  & $0.656 \pm 0.088$ & $0.738 \pm 0.052$ & $0.706 \pm 0.050$ & $0.968 \pm 0.020$ & $0.694 \pm 0.055$ \\
CatBoost      & $\mathbf{0.900 \pm 0.048}$ & $\mathbf{0.849 \pm 0.041}$ & $\mathbf{0.864 \pm 0.020}$ & $0.973 \pm 0.016$ & $\mathbf{0.873 \pm 0.016}$ \\
XGBoost       & $\mathbf{0.926 \pm 0.037}$ & $\mathbf{0.853 \pm 0.038}$ & $\mathbf{0.875 \pm 0.027}$ & $\mathbf{0.979 \pm 0.011}$ & $\mathbf{0.888 \pm 0.025}$ \\
\bottomrule
\end{tabular}%
}

\label{tab:monte_carlo_results_10}
\end{table}
\section{Conclusion}
In this study, we proposed \textit{Quantile-TabDDPM}, a diffusion-based generative framework for numerical tabular data that incorporates \textit{quantile regularization} into the denoising objective. Experimental results on the ECC dataset demonstrate improved fraud detection performance over existing augmentation methods across multiple tree-based classifiers.

These findings highlight the effectiveness of quantile regularization for diffusion-based generative models in highly imbalanced settings. Future work will extend Quantile-TabDDPM to heterogeneous tabular datasets containing both numerical and categorical features.





\section*{Acknowledgements}

This research was funded by the Government of India through the Ministry of Electronics and Information Technology (MeitY) and the Department of Science and Technology (DST).

\section*{Declarations}

\textbf{Competing Interests.} The authors have no competing interests to declare that are relevant to the content of this article.

%
%
%
\bibliographystyle{splncs04}
\bibliography{reference}
\clearpage

\appendix
\section{Quantile-TabDDPM Pipeline}
\label{app:flow}

\begin{figure}[H]

    \centering
    \includegraphics[width=5.2in]{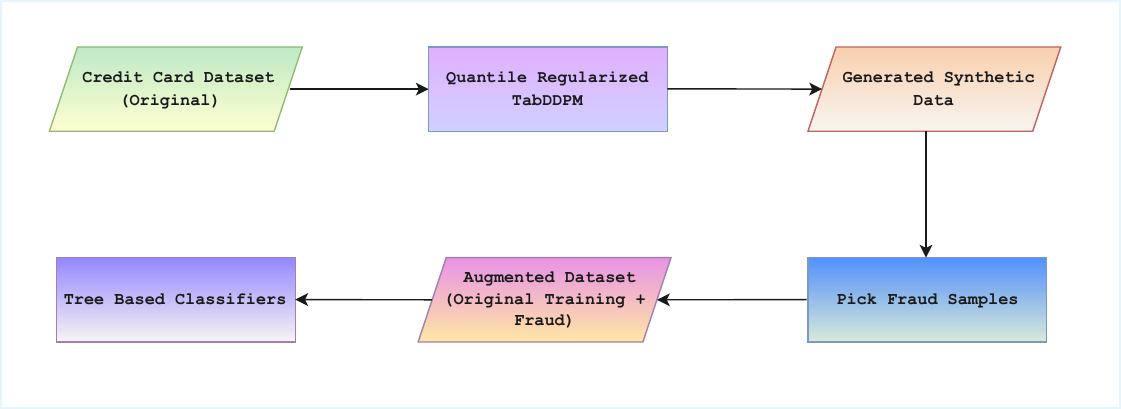}
    \caption{End-to-end workflow of the proposed Quantile-TabDDPM framework for highly imbalanced classification. Rectangles denote processing steps (e.g., training Quantile-TabDDPM, data augmentation, classification), parallelograms represent input/output datasets (original and synthetic data), and arrows indicate the sequential data flow. The process begins with the original credit card dataset, followed by synthetic data generation using TabDDPM with quantile loss, selective extraction of minority (fraud) samples, augmentation of the training dataset, and finally classification using tree-based models.}
    \label{fig:flow-diagram}
\end{figure}
\section{Additional Experimental Details}
\label{app:experimental}

\begin{table}[H]
\centering
\caption{Monte Carlo simulation aggregate results across 30 independent runs for Quantile-TabDDPM with a 90:10 non-fraud-to-fraud ratio.}
\scriptsize 
\setlength{\tabcolsep}{3pt} 
\resizebox{\textwidth}{!}{%
\begin{tabular}{lccccc}
\toprule
Classifier & Precision & Recall & F1-score & ROC-AUC & MCC \\
\midrule
AdaBoost      & $0.818 \pm 0.059$ & $0.787 \pm 0.054$ & $0.796 \pm 0.042$ & $\mathbf{0.973 \pm 0.017}$ & $0.801 \pm 0.041$ \\
RandomForest  & $\mathbf{0.896 \pm 0.068}$ & $\mathbf{0.838 \pm 0.038}$ & $\mathbf{0.855 \pm 0.032}$ & $\mathbf{0.977 \pm 0.014}$ & $\mathbf{0.866 \pm 0.037}$ \\
EasyEnsemble  & $0.642 \pm 0.086$ & $0.728 \pm 0.059$ & $0.693 \pm 0.046$ & $0.967 \pm 0.019$ & $0.681 \pm 0.050$ \\
CatBoost      & $\mathbf{0.902 \pm 0.041}$ & $\mathbf{0.825 \pm 0.048}$ & $\mathbf{0.848 \pm 0.033}$ & $0.971 \pm 0.015$ & $\mathbf{0.862 \pm 0.029}$ \\
XGBoost       & $\mathbf{0.925 \pm 0.039}$ & $\mathbf{0.830 \pm 0.049}$ & $\mathbf{0.859 \pm 0.038}$ & $\mathbf{0.976 \pm 0.014}$ & $\mathbf{0.876 \pm 0.034}$ \\
\bottomrule
\end{tabular}%
}

\label{tab:monte_carlo_results_30}
\end{table}

To further verify the stability of these findings, we extended the analysis to 30 independent Monte Carlo simulations, with results summarized in Table~\ref{tab:monte_carlo_results_30}. The ranking remains unchanged: XGBoost achieves the highest F1-score ($0.859 \pm 0.038$) and MCC ($0.876 \pm 0.034$), followed by RandomForest and CatBoost, while RandomForest attains the best recall and ROC-AUC. Compared with the 10-run results, the mean scores differ only marginally and the standard deviations remain small, indicating that the performance of Quantile-TabDDPM is robust to the number of simulation runs and not an artifact of a particular data split.

\begin{figure}[H]
\centering
\begin{subfigure}{\textwidth}
    \centering
    \includegraphics[width=0.9\textwidth,height=0.4\textheight,keepaspectratio]{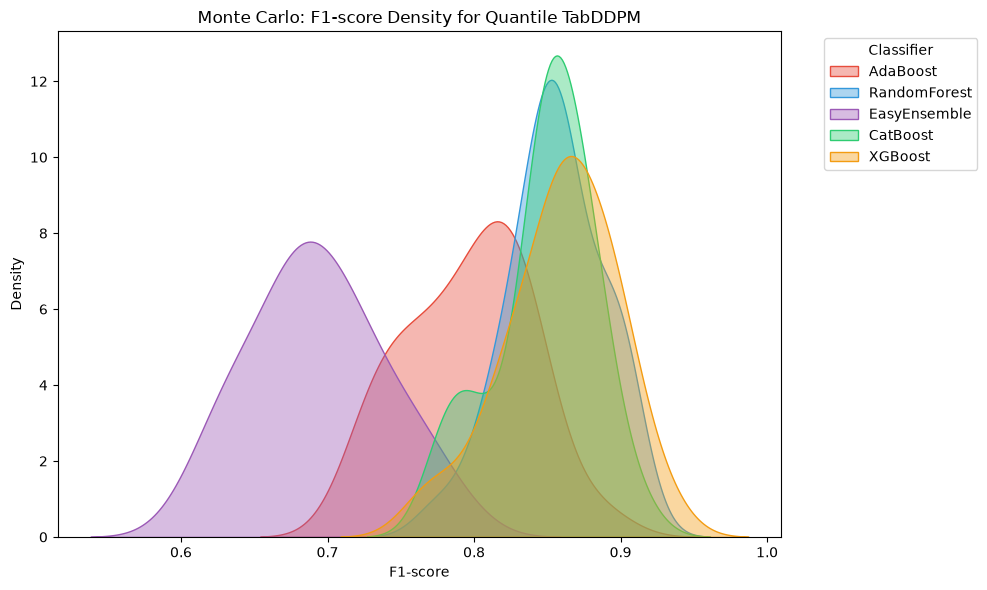}
    \caption{Density plot of F1-score across 30 Monte Carlo simulations.}
    \label{fig:mc30_density_1}
\end{subfigure}

\vspace{4mm}

\begin{subfigure}{\textwidth}
    \centering
    \includegraphics[width=0.9\textwidth,height=0.4\textheight,keepaspectratio]{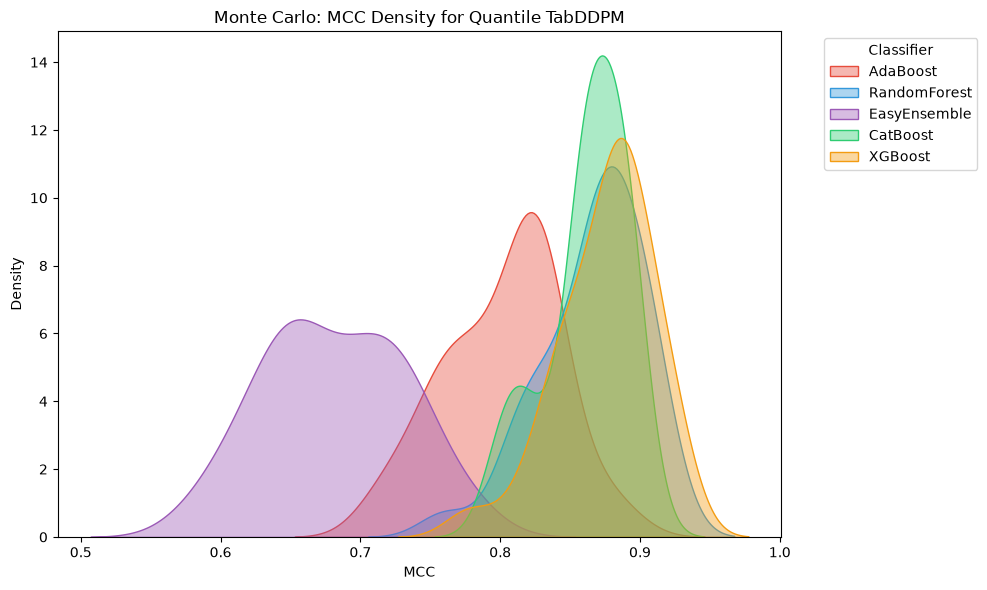}
    \caption{Density plot of MCC across 30 Monte Carlo simulations.}
    \label{fig:mc30_density_2}
\end{subfigure}
\caption{Density plots of classifier performance over 30 independent Monte Carlo simulations for Quantile-TabDDPM (90:10 non-fraud-to-fraud ratio).}
\label{fig:mc30_density}
\end{figure}

Figure~\ref{fig:mc30_density} shows the distributions of F1-score and MCC across the 30 Monte Carlo runs. The densities of XGBoost, RandomForest, and CatBoost are narrow and concentrated at the high end (around 0.85--0.90), with XGBoost's peak slightly further to the right, while AdaBoost centers lower and EasyEnsemble is both the lowest and the most widely spread. These visual trends match the rankings in Table~\ref{tab:monte_carlo_results_30} and further confirm the stability of Quantile-TabDDPM across random splits.

\section{Additional Results}
\label{app:results}

\begin{figure}[H]
    \centering
    \includegraphics[width=1.2\textwidth]{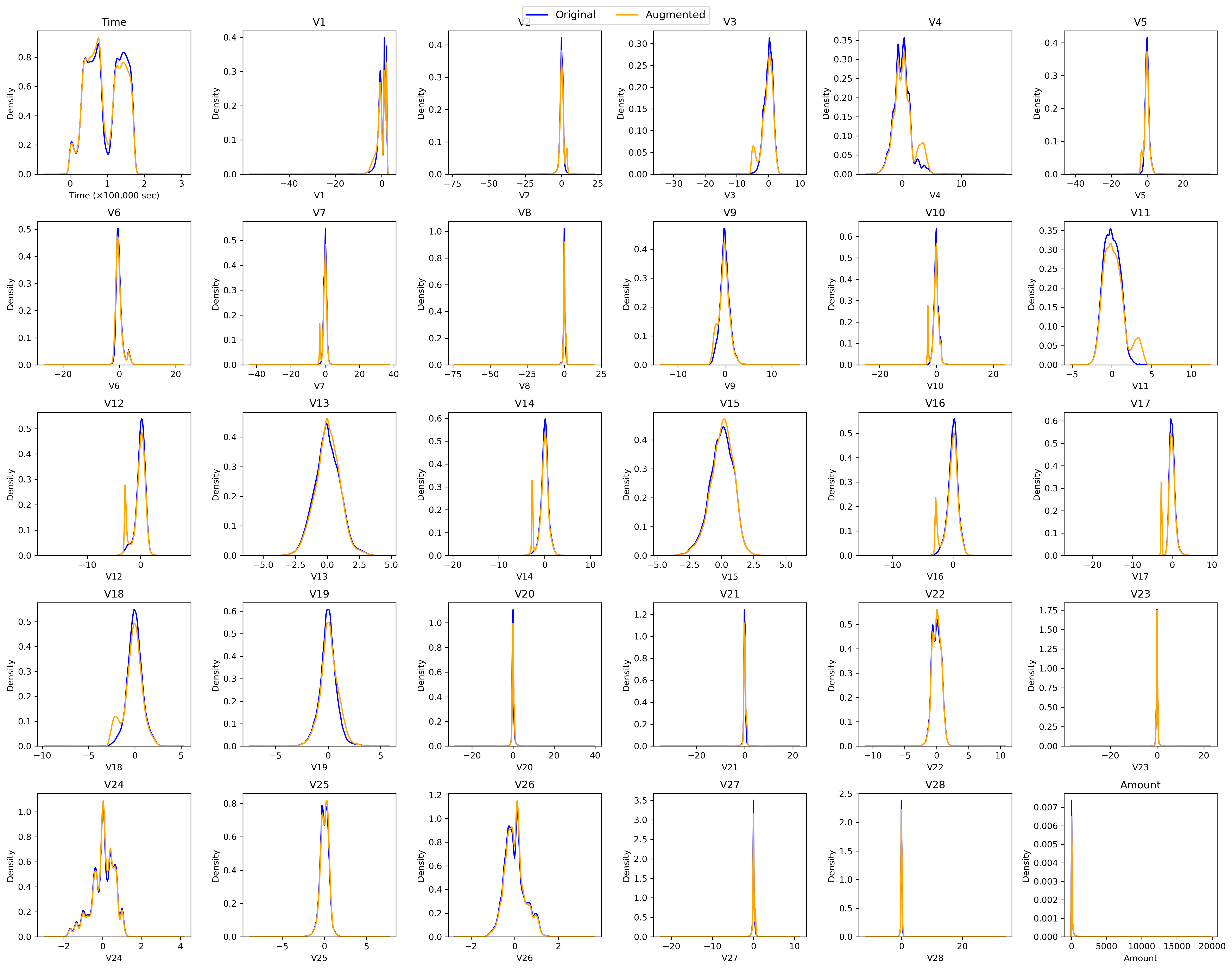}
    \caption{Distribution comparison of original training data vs augmented data using Quantile-TabDDPM}
    \label{fig:dist_quantile_tabddpm}
\end{figure}

\clearpage

\begin{figure}[H]
    \centering
    \includegraphics[width=1.2\textwidth]{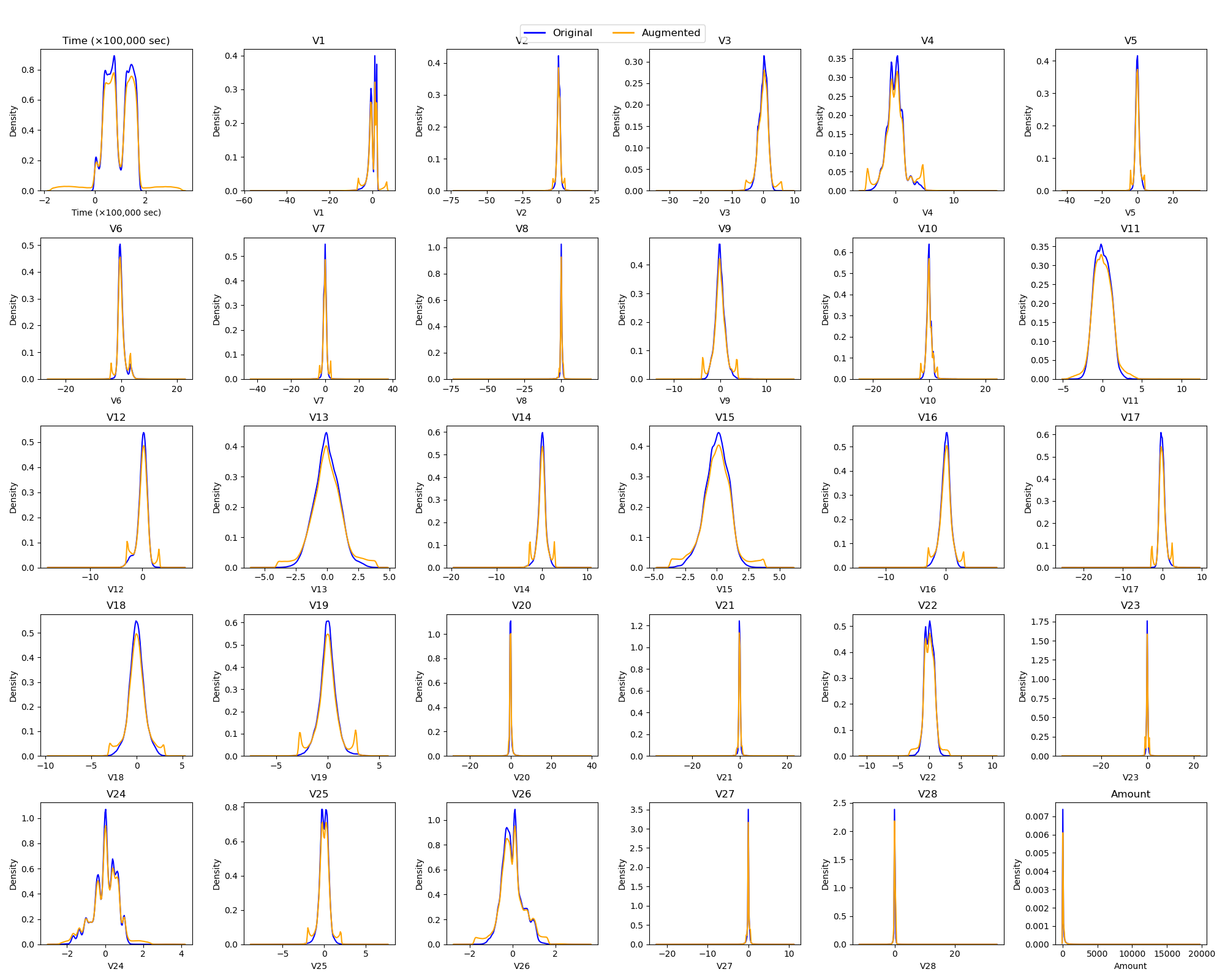}
    \caption{Distribution comparison of original training data vs augmented data using TabDDPM (baseline model).}
    \label{fig:dist_baseline}
\end{figure}

\end{document}